\documentclass[11pt,a4paper]{article}
\usepackage[hyperref]{naaclhlt2018}
\usepackage{times}
\usepackage{latexsym}

\usepackage{url}
\usepackage{multirow}
\usepackage{booktabs,caption}
\usepackage[flushleft]{threeparttable}
\usepackage{subcaption}
\usepackage[nointegrals]{wasysym}
\usepackage{tikz}
\usepackage{amsmath}
\usepackage{amssymb}

\usepackage{todonotes}
\newcommand{\ignore}[1]{}
\newcommand{\event}[1]{\textit{\textbf{#1}}}
\newcommand{\idxevent}[3]{\event{e#1:\textrm{\color{#3}#2}}}
\newcommand{\rel}[1]{{\em #1}}
\newcommand\ph[1]{(\textrm{\color{red}#1?})}
\newcommand{\QN}[1]{#1}
\newcommand{\best}[1]{\underline{#1}}

\newcounter{exctr}
\newcommand{\addexctr}{\refstepcounter{exctr}\theexctr}
\newcounter{eventCtr}
\newcommand{\addeventCtr}{\refstepcounter{eventCtr}\theeventCtr}
\newcommand{\ourkb}{\textsc{TemProb}}
\aclfinalcopy 
\def\aclpaperid{56} 

\ignore{
Improving Time-Aware Tasks via A New Probabilistic Knowledge Base

Improving Time-Aware Tasks via Prior Knowledge Mined from A Large Corpus

Improving Temporal Relation Extraction via A New Probabilistic Knowledge Base

Improving Temporal Relation Extraction via Prior Knowledge Mined from A Large Corpus

A Resource of Temporal Relations for Time-Aware Tasks

A Resource for Temporal Relation Extraction
}

\title{Improving Temporal Relation Extraction with a Globally Acquired \\Statistical Resource}

\author{Qiang Ning,$^1$ Hao Wu,$^2$ Haoruo Peng,$^1$ Dan Roth$^{1,2}$ \\
	Department of Computer Science\\
	$^1$University of Illinois at Urbana-Champaign, Urbana, IL 61801, USA\\
	$^2$University of Pennsylvania, Philadelphia, PA 19104, USA\\
	{\tt \small \{qning2,hpeng7\}@illinois.edu,~\{haowu4,danroth\}@seas.upenn.edu}}

\date{}

\begin{document}
\maketitle
\begin{abstract}
	Extracting temporal relations ({\em before, after, overlapping}, etc.) is a key aspect of understanding events described in natural language. 
	We argue that this task would gain from the availability of a resource that provides prior knowledge in the form of the temporal order that events {\em usually} follow. 
	This paper develops such a resource -- a probabilistic knowledge base acquired in the news domain -- by extracting temporal relations between events from the New York Times (NYT) articles over a 20-year span (1987--2007).
	We show that existing temporal extraction systems can be improved via this resource.
	As a byproduct, we also show that interesting statistics can be retrieved from this resource, which can potentially benefit other time-aware tasks.
	The proposed system and resource are both publicly available\footnote{\url{http://cogcomp.org/page/publication_view/830}}. 
\end{abstract}
\section{Introduction}
Time is an important dimension of knowledge representation. 
In natural language, temporal information is often expressed as relations  between events.
Reasoning over these relations can help figuring out when things happened, estimating how long things take, and summarizing the timeline of a series of events.
Several recent SemEval workshops are a good showcase of the importance of this topic \cite{VGSHKP07,VSCP10,ULADVP13,LCUMAP15,MSAAVMRUK15,BDSPV15,BSCDPV16,BSPP17}.

One of the challenges in temporal relation extraction is  that it requires high-level prior knowledge of the temporal order that events {\em usually} follow. 
In Example~\ref{ex:prior knowledge}, we have deleted events from several snippets from CNN, so that we cannot use our prior knowledge of those events.
\QN{We are also told that \event{e\ref{ev:died}} and \event{e\ref{ev:exploded}} have the same tense, and \event{e\ref{ev:asked}} and \event{e\ref{ev:help}} have the same tense, so we cannot resort to their tenses to tell which one happens earlier.}
As a result, it is very difficult even for humans to figure out the temporal relations (referred to as ``TempRels'' hereafter) between those events. This is because rich temporal information is encoded in the events' names, and this often plays an indispensable role in making our decisions.
\QN{In the first paragraph of Example~\ref{ex:prior knowledge}, it is difficult to understand what really happened without the actual event verbs; let alone the TempRels between them. In the second paragraph, things are even more interesting:}
if we had \idxevent{\ref{ev:asked}}{dislike}{black} and \idxevent{\ref{ev:help}}{stop}{black}, then we would know easily that ``I dislike" occurs {\em after} ``they stop the column".
However, if we had \idxevent{\ref{ev:asked}}{ask}{black} and \idxevent{\ref{ev:help}}{help}{black}, then the relation between \event{e\ref{ev:asked}} and \event{e\ref{ev:help}} is now 
reversed and \event{e\ref{ev:asked}} is {\em before} \event{e\ref{ev:help}}.
\QN{We are in need of the event names to determine the TempRels; however, we do not have them in Example~\ref{ex:prior knowledge}.}
In Example~\ref{ex:prior knowledge 2}, where we show the complete sentences, the task has become much easier for humans due to our prior knowledge, namely, that explosion {\em usually} leads to casualties and that people {\em usually} ask before they get help.
Motivated by these examples (which are in fact very common), we believe in the importance of such a prior knowledge in determining TempRels between events.

\begin{table}[h!]
	\centering\small
	\begin{tabular}{|p{7cm}|}
		\hline
		\textbf{Example~\addexctr\label{ex:prior knowledge}: Difficulty in understanding TempRels when event content is missing.} \QN{Note that \event{e\ref{ev:died}} and \event{e\ref{ev:exploded}} have the same tense, and \event{e\ref{ev:asked}} and \event{e\ref{ev:help}} have the same tense.}\\
		\hline
		More than 10 people have (\idxevent{\addeventCtr\label{ev:died}}{died}{white}), police said. A car (\idxevent{\addeventCtr\label{ev:exploded}}{exploded}{white}) on Friday in the middle of a group of men playing volleyball.
		\\
		\hline
        The first thing I (\idxevent{\addeventCtr\label{ev:asked}}{ask}{white}) is that they (\idxevent{\addeventCtr\label{ev:help}}{help}{white}) writing this column.\\
        \hline
	\end{tabular}
\end{table}

However, most existing systems only make use of rather local features 
of these 
events, which cannot 
represent the prior knowledge  humans have about these events and their ``typical" order. As a result, existing systems almost always attempt to solve the situations shown in Example~\ref{ex:prior knowledge}, even when they are actually presented with input as in Example~\ref{ex:prior knowledge 2}.
The \textbf{first contribution} of this work is thus the construction of such a resource in the form of a probabilistic knowledge base, constructed from a large New York Times (NYT) corpus.
We hereafter name our resource {\em TEMporal relation PRObabilistic knowledge Base} (\ourkb), which can potentially benefit many time-aware tasks. A few example entries of \ourkb~are shown in Table~\ref{tab:format}.
\textbf{Second}, we show that existing TempRel extraction systems can be improved using \ourkb, either in a local method or in a global method (explained later), by a significant margin in performance on the benchmark TimeBank-Dense dataset \cite{CassidyMcChBe14}.
\ignore{
\textbf{Third}, as an example of showing the potential of using \ourkb{}~on other time-aware tasks, we also show significant improvement brought by \ourkb{}~in the cause-effect identification task (on the EventCausality dataset \cite{DoChRo11}).
}

\begin{table}
	\centering\caption{\small \textbf{\ourkb~is a unique source of information of the temporal order that events {\em usually} follow.} The probabilities below do not add up to 100\% because less frequent relations are omitted. The word sense numbers are not shown here for convenience.}
	\label{tab:format}\small
	\begin{tabular}{cc|c|c}
		\hline
		\multicolumn{2}{c|}{Example Pairs}&Before (\%)&After (\%)\\
		\hline
		accept	&		determine	&	42	&	26\\
		ask		&		help		&	86	&	9	\\
		attend	&		schedule	&	1	&	82\\
		accept	&		propose		&	10	&	77\\
		die		&		explode		& 	14	&	83\\
		\multicolumn{4}{c}{\dots}\\
		\hline
	\end{tabular}
\end{table}

\begin{table}[h!]
	\centering\small
	\begin{tabular}{|p{7cm}|}
		\hline
		\textbf{Example~\addexctr\label{ex:prior knowledge 2}: The original sentences in Example~\ref{ex:prior knowledge}.}\\
		\hline
		More than 10 people have (\idxevent{\ref{ev:died}}{died}{black}), police said. A car (\idxevent{\ref{ev:exploded}}{exploded}{black}) on Friday in the middle of a group of men playing volleyball.
		\\
		\hline
		The first thing I (\idxevent{\ref{ev:asked}}{ask}{black}) is that they (\idxevent{\ref{ev:help}}{help}{black}) writing this column.\\
		\hline
	\end{tabular}
\end{table}

The rest of the paper is organized as follows. Section~\ref{sec:related} provides a literature review of TempRels extraction in NLP. Section~\ref{sec:proposed} describes in detail the construction of \ourkb.
In Sec.~\ref{sec:applications}, we show that \ourkb~can be used in existing TempRels extraction systems and lead to significant improvement.
Finally, we conclude in Sec.~\ref{sec:conclusion}.

\section{Related Work}
\label{sec:related}
The TempRels between events can be represented by an edge-labeled graph, where the nodes are events, and the edges are labeled with TempRels \cite{ChambersJu08,DoLuRo12,NingFeRo17}.
Given all the nodes, we work on the TempRel extraction task, which is to assign labels to the edges in a temporal graph (a ``vague'' or ``none'' label is often included to account for the non-existence of an edge).

Early work includes \citet{MVWLP06,ChambersWaJu07,BethardMaKl07,VerhagenPu08}, where the problem was formulated as learning a classification model for determining the label of every edge {\em locally} without referring to other edges (i.e., \underline{local methods}). 
The predicted temporal graphs by these methods may violate the transitive properties that a temporal graph should possess. For example, given three nodes, \event{e1}, \event{e2}, and \event{e3}, a local method can possibly classify (\event{e1},\event{e2})=\rel{before}, (\event{e2},\event{e3})=\rel{before}, and (\event{e1},\event{e3})=\rel{after}, which is obviously wrong since \rel{before} is a transitive relation and (\event{e1},\event{e2})=\rel{before} and (\event{e2},\event{e3})=\rel{before} dictate that (\event{e1},\event{e3})=\rel{before}.
Recent state-of-the-art methods, \cite{ChambersCaMcBe14,MirzaTo16}, circumvented this issue by growing the predicted temporal graph in a multi-step manner, where transitive graph closure is performed on the graph every time a new edge is labeled. This is conceptually solving the structured prediction problem greedily.
Another family of methods resorted to Integer Linear Programming (ILP) \cite{RothYi04} to get exact inference to this problem (i.e., \underline{global methods}), where the entire graph is solved simultaneously  and the transitive properties are enforced naturally via ILP constraints~\cite{BDLB06,ChambersJu08,DenisMu11,DoLuRo12}.
A most recent work brought this idea even further, by incorporating structural constraints into the learning phase as well \cite{NingFeRo17}.

The TempRel extraction task has a strong dependency on prior knowledge, as shown in our earlier examples. However, very limited attention has been paid to generating such a resource and to make use of it; to our knowledge, the \ourkb~proposed in this work is completely new.
We find that the {\em time-sensitive relations} proposed in \citet{JLGSCLS16} is a close one in literature (although it is still very different).
\citet{JLGSCLS16} worked on the knowledge graph completion task.
Based on YAGO2 \cite{HSBW13} and Freebase \cite{BEPST08}, it manually selects a small number of relations that are time-sensitive (10 relations from YAGO2 and 87 relations from Freebase, respectively). Exemplar relations are \textsf{wasBornIn}$\to$\textsf{diedIn}$\to$ and \textsf{graduateFrom}$\to$\textsf{workAt}, where $\to$ means temporally before.

Our work significantly differs from the time-sensitive relations in \citet{JLGSCLS16} in the following aspects. 
First, scale difference: \citet{JLGSCLS16} can only extract a small number of relations ($<$100), but we work on general semantic frames (tens of thousands) and the relations between any two of them, which we think has broader applications.
Second, granularity difference: the smallest granularity in \citet{JLGSCLS16} is one year\footnote{We notice that the smallest granularity in Freebase itself is one day, but \citet{JLGSCLS16} only used years.}, i.e., only when two events happened in different years can they know the temporal order of them, but we can handle implicit temporal orders without having to refer to the physical time points of events (i.e., the granularity can be arbitrarily small).
Third, domain difference: while \citet{JLGSCLS16} extracts time-sensitive relations from structured knowledge bases (where events are explicitly anchored to a time point), we extract relations from unstructured natural language text (where the physical time points may not even exist in text). Our task is more general and it allows us to extract much more relations, as reflected by the 1st difference above.

\QN{Another related work is the VerbOcean \cite{ChklovskiPa04}, which extracts temporal relations between pairs of verbs using manually designed lexico-syntactic patterns (there are in total 12 such patterns), in contrast to the automatic extraction method proposed in this work. In addition, the only termporal relation considered in VerbOceans is \rel{before}, while we also consider relations such as \rel{after}, \rel{includes}, \rel{included}, \rel{equal}, and \rel{vague}. As expected, the total numbers of verbs and \rel{before} relations in VerbOcean is about 3K and 4K, respectively, both of which are much smaller than \ourkb{}, which contains 51K verb frames (i.e., disambiguated verbs), 9.2M $(verb1, verb2, relation)$ entries, and up to 80M temporal relations altogether.}

All these differences necessitate the construction of a new resource for TempRel extraction, which we explain below.


\section{\ourkb: A Probabilistic Resource for TempRels}
\label{sec:proposed}
In the TempRel extraction task, people have usually assumed that events are already given.
However, to construct the desired resource, we need to extract events (Sec.~\ref{subsec:event}) and extract TempRels (Sec.~\ref{subsec:system}), from a large, unannotated\footnote{Unannotated with TempRels.} corpus (Sec.~\ref{subsec:corpus}). 
We also show some interesting statistics discovered in \ourkb~that may benefit other tasks (Sec.~\ref{subsec:interesting}).
In the next, we describe each of these elements.

\subsection{Event Extraction}
\label{subsec:event}
Extracting events and the relations between them (e.g., coreference, causality, entailment, and temporal) have long been an active area in the NLP community.
Generally speaking, an event is considered to be an action associated with corresponding participants involved in this action. 
In this work, following \cite{PengRo16,PengSoRo16,SpiliopoulouHoMi17} we consider semantic-frame based events, which can be directly detected via off-the-shelf semantic role labeling (SRL) tools. 
This aligns well with previous works on event detection
\cite{hovy-EtAl:2013:EVENTS,PengSoRo16}.

Depending on the events of interest, the SRL results are often a superset of events and need to be filtered afterwards \cite{SpiliopoulouHoMi17}.
For example, in ERE \cite{SBSRMEWKR15} and Event Nugget Detection~\cite{MYHSBKS15}, events are limited to a set of predefined types (such as ``Business'', ``Conflict'', and ``Justice'');
in the context of TempRels, existing datasets have focused more on predicate verbs rather than nominals\footnote{Some nominal events were indeed annotated in TimeBank \cite{PHSSGSRSDF03}, but their annotation did not align well with modern nominal-SRL methods.} \cite{PHSSGSRSDF03,Graff02,ULADVP13}. 
Therefore, we only look at verb semantic frames in this work due to the difficulty of getting TempRel annotation for nominal events, and we will use ``verb (semantic frames)'' interchangeably with ``events'' hereafter in this paper.

\ignore{
\subsubsection{Abstraction Level}
When building a knowledge graph of TempRels, certain level of abstractions are often preferred to be able to generalize.
For example, given two events, ``Jack is arrested because of robbery'' and ``John is arrested because of robbery'', one question to ask is ``are they the same or different?''.
One may think that they are different due to their difference between arguments (i.e., ``Jack'' vs. ``John''), but an obvious downside is that there are too many entities of different surface forms to account for in a limited dataset; more importantly, ``rob'' leading to ``being arrested'' is likely to be a common pattern in which their subjects play a minor role.
Based on this intuition, we decide to start from the assumption that two events are considered to be in the same category as long as they share the same predicate (in other words, the knowledge base is built upon disambiguated predicates).
As we show later, this assumption works reasonably well.
We are aware that this level of abstraction may not be perfect, and future work can either perform clustering on those predicates to achieve a higher level of abstraction or plug in entity typing to achieve a finer level of abstraction.
Explorations along this approach need further investigation.
}

\subsection{TempRel Extraction}
\label{subsec:system}
Given the events extracted in a given article (i.e., given the nodes in a graph), we next explain how the TempRels  are extracted (that is, the edge labels in the graph).

\subsubsection{Features}
\label{subsubsec:feat}
We adopt the commonly used feature set in TempRel extraction \cite{DoLuRo12,NingFeRo17} and here we simply list them for reproducibility.
For each pair of nodes, the following features are extracted. (i) The part-of-speech (POS) tags from each individual verb and from its neighboring three words. (ii) The distance between them in terms of the number of tokens. (iii) The modal verbs between the event mention (i.e., \textsf{will, would, can, could, may} and \textsf{might}). (iv) The temporal connectives between the event mentions (e.g., \textsf{before, after} and \textsf{since}). (v) Whether the two verbs have a common
synonym from their synsets in WordNet \cite{Fellbaum98}. (vi) Whether the input event mentions have a common derivational form derived from WordNet. (vii) The head word of the preposition phrase that covers each verb, respectively.

\subsubsection{Learning}
\label{subsubsec:learning}
With the features defined above, we need to train a system that can annotate the TempRels in each document.
The TimeBank-Dense dataset (TBDense)  \cite{CassidyMcChBe14} is known to have the best quality in terms of its high density of TempRels and is a benchmark dataset for the TempRel extraction task.
It contains 36 documents from TimeBank \cite{PHSSGSRSDF03} which were re-annotated
using the dense event ordering framework proposed in \cite{CassidyMcChBe14}.
We follow its label set (denoted by $R$) of \rel{before}, \rel{after}, \rel{includes}, \rel{included}, \rel{equal}, and \rel{vague} in this study.

Due to the slight event annotation difference in TBDense, we collect our training data as follows. We first extract all the verb semantic frames from the raw text of TBDense. Then we only keep those semantic frames that are matched to an event in TBDense (about 85\% semantic frames are kept in this stage). By doing so, we can simply use the TempRel annotations provided in TBDense. Hereafter the TBDense dataset used in this paper refers to this 
version unless otherwise specified.

We group the TempRels by the sentence distance of the two events of each relation\footnote{That is, the difference of the appearance order of the sentence(s) containing the two target events.}. Then we use the averaged perceptron algorithm \cite{FreundSc98} implemented in the Illinois LBJava package \cite{RizzoloRo10} to learn from the training data described above. 
Since only relations that have sentence distance 0 or 1 are annotated in TBDense, we will have two classifiers, one for same sentence relations, and one for neighboring sentence relations, respectively.

Note that TBDense was originally split into Train (22 docs), Dev (5 docs), and Test (9 docs). In all subsequent analysis, we combined Train and Dev and we performed 3-fold cross validation on the 27 documents (in total about 10K relations) to tune the parameters in any classifier.

\subsubsection{Inference} When generating \ourkb, we need to process a large number of articles, so we adopt the greedy inference strategy described earlier due to its computational efficiency \cite{ChambersCaMcBe14,MirzaTo16}.
Specifically, we apply the same-sentence relation classifier before the neighboring-sentence relation classifier; whenever a new relation is added in this article, a transitive graph closure is performed immediately. 
By doing this, if an edge is already labeled during the closure phase, it will not be labeled again, so conflicts are avoided.

\subsection{Corpus}
\label{subsec:corpus}
As mentioned earlier, the source corpus on which we are going to construct \ourkb~is comprised of NYT articles from 20 years (1987-2007)\footnote{https://catalog.ldc.upenn.edu/LDC2008T19}.
It contains more than 1 million documents and we extract events and corresponding features from each document using the Illinois Curator package \cite{ClarkeSrSaRo2012} on Amazon Web Services (AWS) Cloud.
In total, we discovered 51K unique verb semantic frames and 80M relations among them in the NYT corpus (15K of the verb frames had more than 20 relations extracted and 9K had more than 100 relations).

\subsection{Interesting Statistics}
\label{subsec:interesting}

We first describe the notations that we are going to use.
We denote the set of all verb semantic frames by $V$.
Let $D_i, i=1,\dots,N$ be the $i$-th document in our corpus, where $N$ is the total number of documents.
Let $G_i = (V_i, E_i)$ be the temporal graph inferred from $D_i$ using the approach described above, where $V_i\subseteq V$ is the set of verbs\slash events extracted in $D_i$ and $E_i=\{(v_m,v_n,r_{mn})\}_{m<n}\subseteq V_i\times V_i \times R$ is the edge set of $D_i$, which is composed of TempRel triplets; specifically, a TempRel triplet $(v_m,v_n,r_{mn})\in E_i$ represents that in document $D_i$, the TempRel between $v_m$ and $v_n$ is $r_{mn}$.
Due to the symmetry in TempRels, we only keep the triplets with $m<n$ in $E_i$. Assuming that the verbs in $V_i$ are ordered by their appearance order in text, then $m<n$ means that in the $i$-th document, $v_m$ appears earlier in text than $v_n$ does.

Given the usual confusion between that one event is {\em temporally before} another and that one event is {\em physically appearing} before another in text, we will refer to temporally before as \textbf{T-Before} and physically before as \textbf{P-Before}. Using this language, for example, $E_i$ only keeps the triplets that $v_m$ is P-Before $v_n$ in $D_i$.

\subsubsection{Extreme cases}
We first show extreme cases that some events are {\em almost always} labeled as T-Before or T-After in the corpus.
Specifically, for each pair of verbs $v_i,v_j\in V$, we define the following ratios:
\begin{equation}\small
\label{eq:prob}
\eta_b = \frac{C(v_i,v_j,before) }{C(v_i,v_j,before)+C(v_i,v_j,after)}, \eta_a = 1-\eta_b,
\end{equation}
where $C(v_i,v_j,r)$ is the count of $v_i$ P-Before $v_j$ with TempRel $r\in R$:
\begin{equation}\small
\label{eq:count}
C(v_i,v_j,r)=\sum_{i=1}^N \sum_{(v_m,v_n,r_{mn})\in E_i}\mathcal{I}_{\{v_m=v_i\&v_n=v_j\&r_{mn}=r\}},
\end{equation}
where $\mathcal{I}_{\{\cdot\}}$ is the indicator function.
Add-one smoothing technique from language modeling is used to avoid divided-by-zero errors.
In Table~\ref{tab:extreme}, we show some event pairs with either $\eta_b>0.9$ (upper part) or $\eta_a>0.9$ (lower part).

We think the examples from Table~\ref{tab:extreme} are intuitively appealing: \event{chop} happens before \event{taste}, \event{clean} happens after \event{contaminate}, etc.
More interestingly, in the lower part of the table, we show pairs in which the physical order is different from the temporal order: for example, when \event{achieve} is P-Before \event{desire}, it is still labeled as T-After in most cases (104 out of 111 times), which is correct intuitively.
In practice, e.g., in the TBDense dataset \cite{CassidyMcChBe14}, roughly 30\%-40\% of the P-Before pairs are T-After.
Therefore, it is important to be able to capture their temporal order rather than simply taking their physical order if one wants to understand the temporal implication of verbs.

\ignore{
The temporal order of the pairs we show in Table~\ref{tab:extreme} are almost deterministic, i.e., either T-Before or T-After with probability larger than 90\%. We understand the rest 10\% (i.e. those \#T-After's in the upper part and \#T-Before's in the lower part) from two aspects: 1) system imperfection (recall that each $G_i$ is of relatively low quality), and 2) complications brought by the difference in frame arguments (e.g., ``Jack is arrested'' is definitely possible to be T-After ``John is charged'').

Note that only pairs of $\eta_b$ or $\eta_a>0.9$ are shown in Table~\ref{tab:extreme}. Another usage of \ourkb~ is that $\eta_b$ and $\eta_a$ can serve as a soft-decision and be incorporated in subsequent systems, which is exactly the kind of prior knowledge that we have expected.
}

\begin{table}
	\centering\caption{\small \textbf{Several extreme cases from  \ourkb{}}, where some event is almost always labeled to be T-Before or T-After throughout the NYT corpus. By ``extreme'', we mean that either the probability of T-Before or T-After is larger than 90\%. The upper part of the table shows the pairs that are both P-Before and T-Before, while the lower part shows the pairs that are P-Before but T-After. \QN{In \ourkb{}, there are about 7K event pairs being extreme cases.}}
	\label{tab:extreme}\small
	\begin{tabular}{cc|c|c}
		\hline
		\multicolumn{2}{c|}{Example Pairs}&\#T-Before&\#T-After\\
		\hline
		chop.01		&	taste.01	&	133	&	8\\
		concern.01	&	protect.01	&	110	&	10\\
		conspire.01	&	kill.01		&	113	&	6\\
		debate.01	&	vote.01		&	48	&	5\\
		dedicate.01	&	promote.02	&	67	&	7\\
		fight.01	&	overthrow.01&	98	&	8\\
		\hline
		achieve.01	&	desire.01	&	7	&	104	\\
		admire.01	&	respect.01	&	7	&	121\\
		clean.02	&	contaminate.01&	3	&	82\\
		defend.01	&	accuse.01	&	13	&	160\\
		die.01		&	crash.01	&	8	&	223\\
		overthrow.01&	elect.01	&	3	&	100\\
		\hline
	\end{tabular}
\end{table}

\begin{figure*}
	\centering
	\begin{subfigure}[htbp!]{0.49\textwidth}
		\centering
		\includegraphics[width=\textwidth]{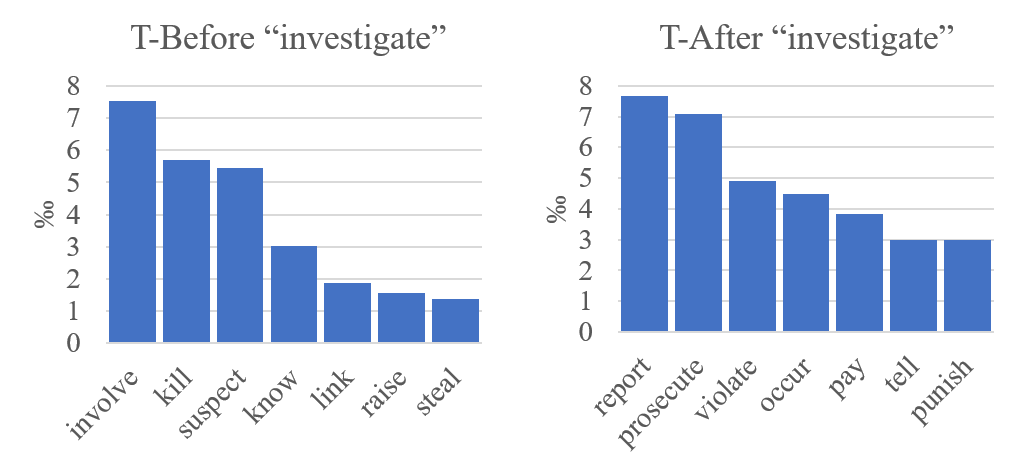}
		\caption{investigate}
	\end{subfigure}%
	~
	\begin{subfigure}[htbp!]{0.49\textwidth}
		\centering
		\includegraphics[width=\textwidth]{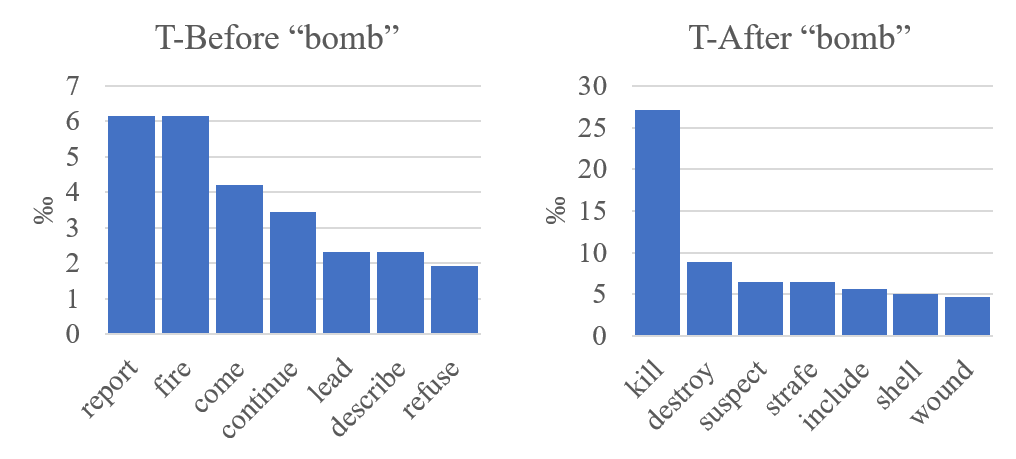}
		\caption{bomb}
	\end{subfigure}
	
	\begin{subfigure}[htbp!]{0.49\textwidth}
		\centering
		\includegraphics[width=\textwidth]{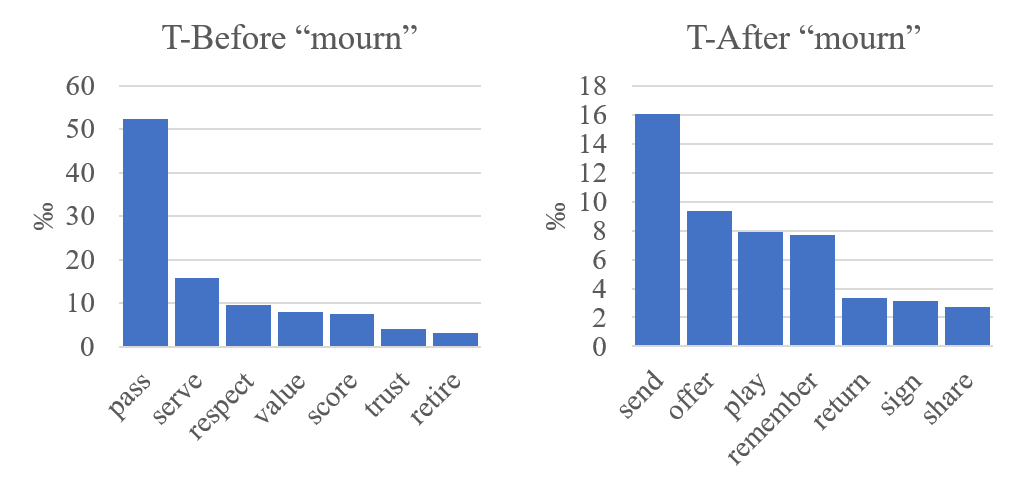}
		\caption{mourn}
	\end{subfigure}
	~
	\begin{subfigure}[htbp!]{0.49\textwidth}
		\centering
		\includegraphics[width=\textwidth]{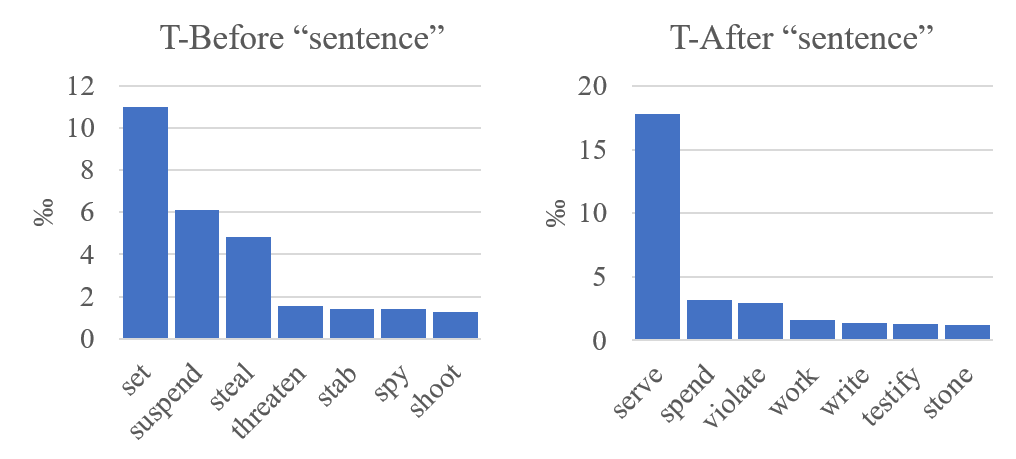}
		\caption{sentence}
	\end{subfigure}
	\caption{\small \textbf{Top events that most frequently precede or follow ``investigate'', ``bomb'', ``mourn'', or ``sentence'' in time}, sorted by their conditional probabilities in \permil. Word senses have been disambiguated and the ``bomb'' and ``sentence'' here are their verb meanings. There are some possible errors (e.g., \event{report} is T-Before \event{bomb}) and some unclear pairs (e.g., \event{know} is T-Before \event{investigate} and \event{play} is T-After \event{mourn}), but overall the event sequences discovered here are reasonable. More examples can be found in the appendix.}
	\label{fig:distribution}
\end{figure*}

\subsubsection{Distribution of Following Events}
\label{subsubsec:distribution}
\ignore{In addition to the extreme cases shown in Table~\ref{tab:extreme}, we also show analysis of the distribution of preceding and following events in this section.}

For each verb $v$, we define the marginal count of $v$ being P-Before to arbitrary verbs with TempRel $r\in R$ as $C(v,r)=\sum_{v_i\in V}C(v,v_i,r)$.
Then for every other verb $v^\prime$, we define
\begin{equation}\small
\label{eq:cond prob before}
P(v~\textrm{T-Before}~v^\prime \vert v~\textrm{T-Before}) \triangleq \frac{C(v,v^\prime,before)}{C(v,before)},
\end{equation}
which is the probability of $v$ T-Before $v^\prime$, conditioned on $v$ T-Before anything.
Similarly, we define
\begin{equation}\small
\label{eq:cond prob after}
P(v~\textrm{T-After}~v^\prime \vert v~\textrm{T-After}) \triangleq \frac{C(v,v^\prime,after)}{C(v,after)}.
\end{equation}
For a specific verb, e.g., $v$=\event{investigate}, each verb $v^\prime\in V$ is sorted by the two conditional probabilities above. Then the most probable verbs that temporally precede or follow $v$ are shown in Fig.~\ref{fig:distribution}, where the y-axes are the corresponding conditional probabilities.
We can see reasonable event sequences like \{\event{involve}, \event{kill}, \event{suspect}, \event{steal}\}$\to$\event{investigate}$\to$\{\event{report}, \event{prosecute}, \event{pay}, \event{punish}\}, which indicates the possibility of using \ourkb~for event sequence predictions or story cloze tasks.
There are also suspicious pairs like \event{know} in the T-Before list of \event{investigate} (Fig.~\ref{fig:distribution}a), \event{report} in the T-Before list of \event{bomb} (Fig.~\ref{fig:distribution}b), and \event{play} in the T-After list of \event{mourn} (Fig.~\ref{fig:distribution}c). Since the arguments of these verb frames are not considered here, whether these few seemingly counter-intuitive pairs come from system error or from a special context needs further investigation.

\ignore{Now we have seen some interesting examples when we aggregate information from \ourkb. In the next section, we will show quantitative analyses of how \ourkb{}~ can help other tasks effectively.}

\section{Experiments}
\label{sec:applications}
In the above, we have explained the construction of \ourkb~and shown some interesting examples from it, which were meant to visualize its correctness.
In this section, we first quantify the correctness of the prior obtained in \ourkb, and then show \ourkb~can be used to improve existing TempRel extraction systems.

\subsection{Quality Analysis of \ourkb}
\label{subsec:quality}
In Table~\ref{tab:extreme}, we showed examples with either $\eta_b$ or $\eta_a>0.9$. We argued that they {\em seem} correct. Here we quantify the ``correctness'' of $\eta_b$ and $\eta_a$ based on TBDense.
Specifically, we collected all the gold T-Before and T-After pairs.
Let $\tau\in[0.5,1)$ be a constant threshold. 
Imagine a naive predictor \QN{such} that for each pair of events $v_i$ and $v_j$, if $\eta_b>\tau$, it predicts that $v_i$ is T-Before $v_j$; if $\eta_a>\tau$, it predicts that $v_i$ is T-After $v_j$; otherwise, it predicts \QN{that} $v_i$ is T-Vague to $v_j$.
We expect that a higher $\eta_b$ (or $\eta_a$) represents a higher confidence for an instance to be labeled T-Before (or T-After).

\begin{table}[htbp!]
	\centering
	\caption{\small \textbf{Validating $\eta_b$ and $\eta_a$ from \ourkb}~based on the T-Before and T-After examples in TBDense. Performances are decomposed into same sentence examples (Dist=0) and contiguous sentence examples (Dist=1). A larger threshold leads to a higher precision, so $\eta_b$ and $\eta_a$ indeed represent a notion of confidence.}
	\label{tab:thresholding}\small
	\begin{tabular}{ c|cc|cc } 
		\hline
		\multirow{2}{*}{Threshold $\tau$}&\multicolumn{2}{c|}{Dist=0}&\multicolumn{2}{c}{Dist=1}\\\cline{2-5}&P&R&P&R\\
		\hline
		0.5 	& 	65.6	&	61.3	 	& 	58.5	&	53.3	\\
		0.6	&	69.8	&	44.5		&	60.5	&	36.9		\\
		0.7	&	74.6	&	29.2		&	63.6	&	18.7		\\
		0.8	&	81.0	&	13.9		&	64.8	&	6.9		\\
		0.9	&	\best{82.9}	&	5.0		&	\best{76.9}	&	1.2	\\
		\hline
	\end{tabular}
\end{table}

Table~\ref{tab:thresholding} shows the performance of this predictor, which meets our expectation and thus justifies the validity of \ourkb. 
As we gradually increase the value of $\tau$ in Table~\ref{tab:thresholding}, the precision increases in roughly the same pace with $\tau$, which indicates that the values of $\eta_b$ and $\eta_a$\footnote{Recall the definitions of $\eta_b$ and $\eta_a$ in Eq.~\eqref{eq:prob}.} from \ourkb~indeed represent the confidence level. 
The decrease in recall is also expected because more examples are labeled as T-Vague when $\tau$ is larger. 
\ignore{
Specifically, when $\tau=0.9$, the precision for same sentence pairs is even comparable to the manually designed syntactic rules in CAEVO \cite{ChambersCaMcBe14}, a state-of-the-art system, only with a better recall (refer to Table~4 therein). 
This observation motivates us to add the prior distributions as regularization terms in inference, as we show later in Sec.~\ref{subsubsec:reg}.
}

To further justify the quality, we also used another dataset that is not in the TempRel domain. Instead, we downloaded the EventCausality dataset\footnote{\url{http://cogcomp.org/page/resource_view/27}} \cite{DoChRo11}. For each causally related pair \event{e1} and \event{e2}, if EventCausality annotates that \event{e1} causes \event{e2}, we changed it to be T-Before; if EventCausality annotates that \event{e1} is caused by \event{e2}, we changed it to be T-after. Therefore, based on the assumption that the cause event is T-Before the result event, we converted the EventCausality dataset to be a TempRel dataset and it thus could also be used to evaluate the quality of \ourkb.
We adopted the same predictor used in Table~\ref{tab:thresholding} with $\tau=0.5$ and in Table~\ref{tab:causal}, we compared it with two baselines: (i) always predicting T-Before and (ii) always predicting T-After.
First, the accuracy (66.2\%) in Table~\ref{tab:causal} is rather consistent with its counterpart in Table~\ref{tab:thresholding}, confirming the stability of statistics from \ourkb.
Second, by directly using the prior statistics $\eta_b$ and $\eta_a$ from \ourkb, we can improve the precision of both labels with a significant margin \QN{relative to the two baselines} (17.0\% for ``T-Before'' and 15.9\% for ``T-After'').  Overall, the accuracy was improved by 11.5\%.

\begin{table}[htbp!]
	\centering
	\caption{\small {\bf Further justification of $\eta_b$ and $\eta_a$ from \ourkb~on the EventCausality dataset}. The thresholding predictor from Table~\ref{tab:thresholding} with $\tau=0.5$ is used here. Compared to always predicting the majority label (i.e., T-Before in this case), $\tau=0.5$ significantly improved the performance for both labels, with the overall accuracy improved by 11.5\%.}
	\label{tab:causal}
	\small
	\begin{tabular}{ c|cc|cc|c } 
		\hline
		\multirow{2}{*}{System}&\multicolumn{2}{c|}{T-Before}&\multicolumn{2}{c|}{T-After}&\multirow{2}{*}{Acc.}\\\cline{2-5}&P&R&P&R\\
		\hline
		T-Before Only&54.7&100.0&0&0&54.7\\
		T-After Only&0&0&45.3&100&45.3\\
		$\tau=0.5$&\best{71.7}&63.3&\best{61.2}&69.8&\best{66.2}\\
		\hline
	\end{tabular}
\end{table}

\subsection{Improving TempRel Extraction}
\label{subsec:temporal}
The original purpose of \ourkb~was to improve TempRel extraction. 
We show it from two perspectives: How effective the prior distributions obtained from \ourkb~are (i) as features in local methods and (ii) as regularization terms in global methods.
The results below were evaluated on the test split of TB-Dense \cite{CassidyMcChBe14}.

\subsubsection{Improving Local Methods}
\label{subsubsec:use as feature}
We first test how well the prior distributions from  \ourkb{} can be used as features in improving local methods for TempRel extraction.
In Table~\ref{tab:use prior locally}, we used the original feature set proposed in Sec.~\ref{subsubsec:feat} as the baseline, and added the prior distribution obtained from \ourkb{} on top of it. Specifically, we added $\eta_b$ (see Eq.~\eqref{eq:prob}) and $\{f_r\}_{r\in R}$, respectively, where $\{f_r\}_{r\in R}$ is the prior distributions of all labels, i.e.,
\begin{equation}
\label{eq:all label prior}
f_r(v_i,v_j)=\frac{C(v_i,v_j,r)}{\sum_{r^\prime\in R}{C(v_i,v_j,r^\prime)}},~r\in R.
\end{equation}
Recall function $C$ is defined in Eq.~\eqref{eq:count}.
All comparisons were decomposed to same sentence relations (Dist=0) and neighboring sentence relations (Dist=1) for a better understanding of the behavior.
All classifiers were trained using the averaged perceptron algorithm \cite{FreundSc98} and tuned by 3-fold cross validation.

From Table~\ref{tab:use prior locally}, we can see that simply adding $\eta_b$ into the feature set could improve the original system F$_1$ by 1.8\% (Dist=0) and 3.0\% (Dist=1).
If we further add as features the full set of prior distributions $\{f_r\}_{r\in R}$, the improvement comes to 2.7\% and 6.5\%, respectively. 
Noticing that the feature is more helpful for Dist=1, we think that it is because distant pairs usually have less lexical dependency and thus need more prior information provided by our new feature.
\QN{With Dist=0 and Dist=1 combined (numbers not shown in the Table), the 3rd line improved the "original" by 4.7\% in F$_1$ and by 5.1\% in the temporal awareness F-score (another metric used in the TempEval3 workshop).}

\begin{table}[htbp!]
	\centering
	\caption{\small \textbf{Using prior distributions derived from \ourkb~as features in an example local method.} Incorporating $\eta_b$ to the original feature set already yields better performance. By using the full set of prior distributions, $\{f_r\}_{r\in R}$, the final system improves the original in almost all metrics, \QN{and the improvement is statistically significant with p$<$0.005 per the McNemar's test.}}
	\label{tab:use prior locally}\small
	\begin{tabular}{ c|c|c|c|c|c|c } 
		\hline
		\multirow{2}{*}{Feature Set}&\multicolumn{3}{c|}{Dist=0}&\multicolumn{3}{c}{Dist=1}\\\cline{2-7}&P&R&F$_1$&P&R&F$_1$\\
		\hline
		Original		&	44.5	&	{57.1}	&	50.0	&	49.0	&	36.9	&	42.1\\
		+$\eta_b$&	{46.2}	&	{58.9}	&	{51.8}	&	\best{55.3}	&	38.1	&	{45.1}\\
		+$\{f_r\}_{r\in R}$&	\best{46.9}	&	\best{60.1}	&	\best{52.7}	&	{51.3}	&	\best{46.2}	&	\best{48.6}\\
		\hline
	\end{tabular}
	
	\begin{tablenotes}
		\item \textbf{Note} The performances here are consistently lower than those in Table~\ref{tab:thresholding} because in Table~\ref{tab:thresholding}, only T-Before and T-After examples are considered, but here all labels are taken into account and the problem is more practical and harder.
	\end{tablenotes}
\end{table}

\subsubsection{Improving Global Methods}
\label{subsubsec:reg}
As mentioned earlier in Sec.~\ref{sec:related}, many systems adopt a global inference method via integer linear programming (ILP) \cite{RothYi04} to enforce transitivity constraints over an entire temporal graph \cite{BDLB06,ChambersJu08,DenisMu11, DoLuRo12, NingFeRo17}.
In addition to the usage shown in Sec.~\ref{subsubsec:use as feature}, the prior distributions from \ourkb~can also be used to regularize the conventional ILP formulation.
Specifically, in each document, let $\mathcal{I}_r(ij)\in\{0,1\}$ be the indicator function of relation $r$ for event $i$ and event $j$; let $x_r(ij)\in[0,1]$ be the corresponding soft-max score obtained from the local classifiers (depending on the sentence distance between $i$ and $j$).
Then the ILP objective for global inference is formulated as follows.
\begingroup\makeatletter\def\f@size{10}\check@mathfonts
\def\maketag@@@#1{\hbox{\m@th\large\normalfont#1}}%
\begin{align}\small
\label{eq:ilp}
\hat{\mathcal{I}} = \textrm{arg}\underset{\mathcal{I}}{\textrm{max}}\sum_{ij\in\mathcal{E}}\sum_{r\in R} (x_r(ij)+\underline{\lambda f_r(ij)}) \mathcal{I}_r(ij)\\
\textrm{s.t.}\quad\underset{\textrm{(uniqueness)}}{\Sigma_{r}{\mathcal{I}_r(ij)} = 1},~\underset{\textrm{(symmetry)}}{\mathcal{I}_r(ij) = \mathcal{I}_{\bar{r}}(ji),}\nonumber\\
\underset{\textrm{(transitivity)}}{\mathcal{I}_{r_1}(ij)+\mathcal{I}_{r_2}(jk)-\Sigma_{m=1}^M \mathcal{I}_{r_{3}^m}(ik)\le 1,}\nonumber
\end{align}\endgroup
for all distinct events $i$, $j$, and $k$, where $\mathcal{E}=\{ij~|~\textrm{sentence dist$(i,j)$$\le 1$}\}$, $\lambda$ adjusts the regularization term and was heuristically set to 0.5 in this work, $\bar{r}$ is the reverse relation of $r$, and $M$ is the number of possible relations for $r_{3}$ when $r_1$ and $r_2$ are true.
Note our difference from the ILP in \cite{NingFeRo17} is the underlined regularization term $f_r(ij)$ (which itself is defined in Eq.~\eqref{eq:all label prior}) obtained from \ourkb.

\begin{table}[htbp!]
\centering
\caption{\small \textbf{Regularizing global methods by the prior distribution derived from \ourkb.} The ``+'' means adding a component on top of its preceding line. F$_{\text{aware}}$ is the temporal awareness F-score, another evaluation metric used in TempEval3. The baseline system is to use (unregularized) ILP on top of the original system in Table~\ref{tab:use prior locally}. System~3 is the proposed. \QN{Per the McNemar's test, System~3 is significantly better than System~1 with p$<$0.0005.}}
\small
\begin{tabular}{c|c|c|c|c|c}
\hline
No.&System		&	P		&	R		&	F$_1$	&	F$_{\text{aware}}$\\
\hline
1&Baseline
			&	48.1	&	44.4	&	46.2	&	42.5\\	
2&+Feature: $\{f_r\}_{r\in R}$
			&	50.6	&	52.0	&	51.3	&	49.1\\
3&+Regularization
			&	\best{51.3}	&	\best{53.0}	&	\best{52.1}	&	\best{49.6}\\
\hline
\end{tabular}
\label{tab:regularization}
\end{table}

We present our results on the test split of TB-Dense in Table~\ref{tab:regularization}, which is an ablation study showing step-by-step improvements in two metrics.
In addition to the straightforward precision, recall, and F$_1$ metric, we also compared the F$_1$ of the temporal awareness metric used in TempEval3 \cite{ULADVP13}.
The awareness metric performs graph reduction and closure before evaluation so as to better capture how useful a temporal graph is. Details of this metric can be found in \citet{UzzamanAllen11,ULADVP13,NingFeRo17}.

\begin{table}[htbp!]
	\centering
	\small
	\caption{\small \QN{Label-wise performance improvement of System~3 over System~1 in Table~\ref{tab:regularization}. We can see that incorporating \ourkb{} improves the recall of \rel{before} and \rel{after}, and improves the precision of all labels, with a slight drop in the recall of \rel{vague}.}}
	\begin{tabular}{c|c|c|c}
		\hline
		Label&	P	&	R	&	F$_1$\\
		\hline
		before&+0.3&+15&+6\\
		after&+4&+4&+4\\
		equal&+11&0&+2\\
		includes&+17&0&+0.2\\
		included&+8&0&+2\\
		vague&+3&-4&-1\\
		\hline
	\end{tabular}
	\label{tab:decompose}
\end{table}

In Table~\ref{tab:regularization}, the baseline used the original feature set proposed in Sec.~\ref{subsubsec:feat} and applied global ILP inference with transitivity constraints.
Technically, it is to solve Eq.~\eqref{eq:ilp} with $\lambda=0$ (i.e., unregularized) on top of the original system in Table~\ref{tab:use prior locally}.
Apart from some implementation details, this baseline is also the same as many existing global methods as \citet{ChambersJu08,DoLuRo12}.
System~2, ``+Feature: $\{f_r\}_{r\in R}$'', is to add prior distributions as features when training the local classifiers. Technically, the scores $x_r(ij)$'s in Eq.~\eqref{eq:ilp} used by baseline were changed. We know from Table~\ref{tab:use prior locally} that adding $\{f_r\}_{r\in R}$ made the local decisions better.
Here the performance of System~2 shows that this was also the case for the global decisions made via ILP: both precision and recall got improved, and F$_1$ and awareness were both improved by a large margin, with 5.1\% in F$_1$ and 6.6\% in awareness F$_1$.
On top of this, System~3 sets $\lambda=0.5$ in Eq.~\eqref{eq:ilp} to add regularizations to the conventional ILP formulation. The sum of these regularization terms represents a confidence score of how coherent the predicted temporal graph is to our \ourkb, which we also want to maximize.
Even though a considerable amount of information from \ourkb{} had already been encoded as features (as shown by the large improvements by System~2), these regularizations were still able to further improve the precision, recall and awareness scores.
To sum up, the total improvement over the baseline system brought by \ourkb~is 5.9\% in F$_1$ and 7.1\% in awareness F$_1$, both with a notable margin.
\QN{Table~\ref{tab:decompose} furthermore decomposes this improvement into each TempRel label.}

\begin{table}[htbp!]
\centering
\caption{\small \textbf{Comparison of the proposed TempRel extraction method with two best-so-far systems in terms of two metrics.} Since \ourkb~is only on verb events extracted by SRL, {\em Partial} TBDense is the focus of our work, where we can see significant improvement brought by simply using the prior knowledge from \ourkb. \QN{Per the McNemar's test, Line~3 is better than Line~2 with p$<$0.0005}. For interested readers, we also naively augmented the proposed method to the {\em complete} TBDense and show state-of-the-art performance on it.}
\small
\begin{tabular}{c|c|c|c|c|c}
\hline
No.&System		&	P		&	R		&	F$_1$	&	F$_{\text{aware}}$\\
\hline
\multicolumn{6}{c}{\em Partial TBDense*: Focus of this work.}\\
\hline
1&CAEVO
&	\best{52.3}	&	43.7	&	47.6	&	46.7\\
2&\citet{NingFeRo17}
&	47.4	&	56.3	&	51.5	&	49.1\\
3&Proposed	&	50.0	&	\best{62.4}	&	\best{55.5}	&	\best{52.8}\\
\hline
\multicolumn{6}{c}{\em Complete TBDense: Naive augmentation.}\\
\hline
4&CAEVO
&	\best{51.8}	&	32.6	&	40.0	&	45.7\\
5&\citet{NingFeRo17}
&	46.2	&	40.6	&	43.2	&	48.5\\
6&Proposed**
&	47.2	&	\best{42.4}	&	\best{44.7}	&	\best{49.2}\\
\hline
\end{tabular}
\label{tab:compare}

\begin{tablenotes}
\small
\item *Note that \ourkb~is only available for events extracted by SRL (See Sec.~\ref{subsubsec:learning} for details).
\item **Augment the output of Line~3 with predictions from \citet{NingFeRo17}.
\end{tablenotes}
\end{table}

To compare with state-of-the-art systems, which all used gold event properties (i.e., Tense, Aspect, Modality, and Polarity), we retrained System~3 in Table~\ref{tab:regularization} with these gold properties and show the results in Table~\ref{tab:compare}.
We reproduced the results of CAEVO\footnote{\url{https://github.com/nchambers/caevo}} \cite{ChambersCaMcBe14} and \citet{NingFeRo17}\footnote{\url{http://cogcomp.org/page/publication_view/822}} and evaluated them on the partial TBDense test split\footnote{\QN{There are 731 relations in the partial TBDense test split (201 \rel{before}, 138 \rel{after}, 39 \rel{includes}, 31 \rel{included}, 14 \rel{equal}, and 308 \rel{vague}).}}.
Under both metrics, the proposed system achieved the best performance.
An interesting fact is that even without these gold properties, our System~3 in Table~\ref{tab:regularization} was already better than CAEVO (on Line~1) and \citet{NingFeRo17} (on Line~2) in both metrics.
This is appealing because in practice, those gold properties may not exist, but our proposed system can still generate state-of-the-art performance without them.

For readers who are interested in the complete TBDense dataset, we also performed a naive augmentation as follows. Recall that System~3 only makes predictions to a subset of the complete TBDense dataset. We kept this subset of predictions, and filled the missing predictions by \citet{NingFeRo17}.
Performances of this naively augmented proposed system is compared with CAEVO and \citet{NingFeRo17} on the {\em complete} TBDense dataset.
We can see that by replacing with predictions from our proposed system, \citet{NingFeRo17} got a better precision, recall, F$_1$, and awareness F$_1$, which is the new state-of-the-art on all reported performances on this dataset.
Note that the awareness F$_1$ scores on Lines~4-5 are consistent with reported values in \citet{NingFeRo17}.
To our knowledge, the results in Table~\ref{tab:compare} is the first in literature that reports performances in both metrics, and it is promising to see that the proposed method outperformed state-of-the-art methods in both metrics.

\ignore{
The numbers in Table~\ref{tab:regularization} cannot be directly compared with the state-of-the-art system \cite{NingFeRo17} evaluated on the {\em complete} TBDense dataset (recall that we only work on a subset of the events extracted from Verb-SRL). 
For readers interested in the {\em complete} TBDense dataset, we plugged the predictions from the proposed $\spadesuit$ system into the system on line 13, Table~4 in \cite{NingFeRo17}\footnote{The reproduction package of \cite{NingFeRo17} is available at \url{https://github.com/qiangning/StructTempRel-EMNLP17}}, while keeping all its other predictions. Using the metric in \cite{NingFeRo17}, we improved its reported performance from P=45.6, R=51.9, F$_1$=48.5 to P=47.1, R=55.2, F$_1$=50.8, which is the new state-of-the-art to our knowledge.
}
\ignore{
\ph{relation to Quang's causal system, Haoruo's LM}
\ph{how to infer causality from 1M graphs}
}
\section{Conclusion}
\label{sec:conclusion}
Temporal relation (TempRel) extraction is an important and challenging task in NLP, partly due to its strong dependence on prior knowledge.
Motivated by practical examples, this paper argues that a resource of the temporal order that events {\em usually} follow is helpful.
To construct such a resource, we automatically processed a large corpus from NYT with more than 1 million documents using an existing TempRel extraction system and obtained the {\em TEMporal relation PRObabilistic knowledge Base} (\ourkb{}).
The \ourkb{}~is a good showcase of the capability of such prior knowledge, and it has shown its power in improving existing TempRel extraction systems on a benchmark dataset, TBDense.
The resource and the system reported in this paper are both publicly available\footnote{\url{http://cogcomp.org/page/publication_view/830}} and we hope that it can foster more investigations into time-related tasks.

\section*{Acknowledgements}
\QN{
We thank all the reviewers for providing useful comments. This research is supported in part by a grant from the Allen Institute for Artificial Intelligence (allenai.org); the IBM-ILLINOIS Center for Cognitive Computing Systems Research (C3SR) - a research collaboration as part of the IBM AI Horizons Network; by DARPA under agreement number FA8750-13-2-0008; and by the Army Research Laboratory (ARL) under agreement W911NF-09-2-0053.}

\QN{
The U.S. Government is authorized to reproduce and distribute reprints for Governmental purposes notwithstanding any copyright notation thereon. 
The views and conclusions contained herein are those of the authors and should not be interpreted as necessarily representing the official policies or endorsements, either expressed or implied, of DARPA or the U.S. Government.
Any opinions, findings, conclusions or recommendations are those of the authors and do not necessarily reflect the view of the ARL.}

\bibliography{naaclhlt2018_2,kbcom18,cited-long,ccg-long}
\bibliographystyle{acl_natbib.bst}
\onecolumn
\clearpage
\appendix

\section{Supplemental Material}
\label{sec:supplemental}

\begin{figure*}[h!]
	\centering
	\begin{subfigure}[htbp!]{0.49\textwidth}
		\centering
		\includegraphics[width=\textwidth]{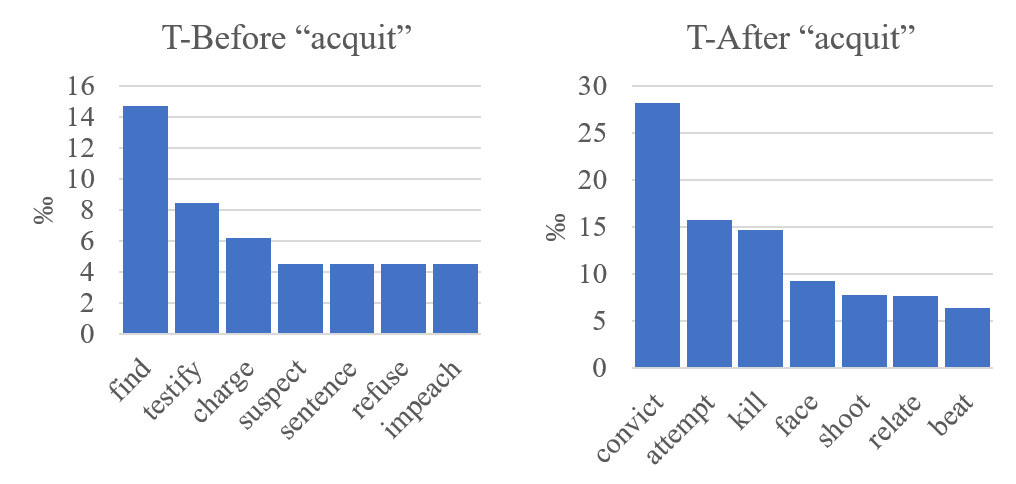}
		\caption{acquit}
	\end{subfigure}%
	~
	\begin{subfigure}[htbp!]{0.49\textwidth}
		\centering
		\includegraphics[width=\textwidth]{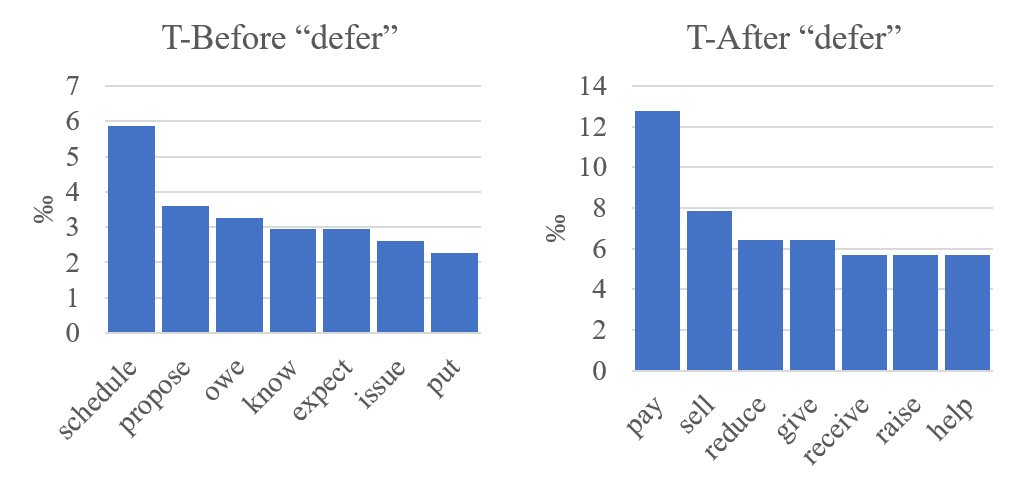}
		\caption{defer}
	\end{subfigure}
	
	\begin{subfigure}[htbp!]{0.49\textwidth}
		\centering
		\includegraphics[width=\textwidth]{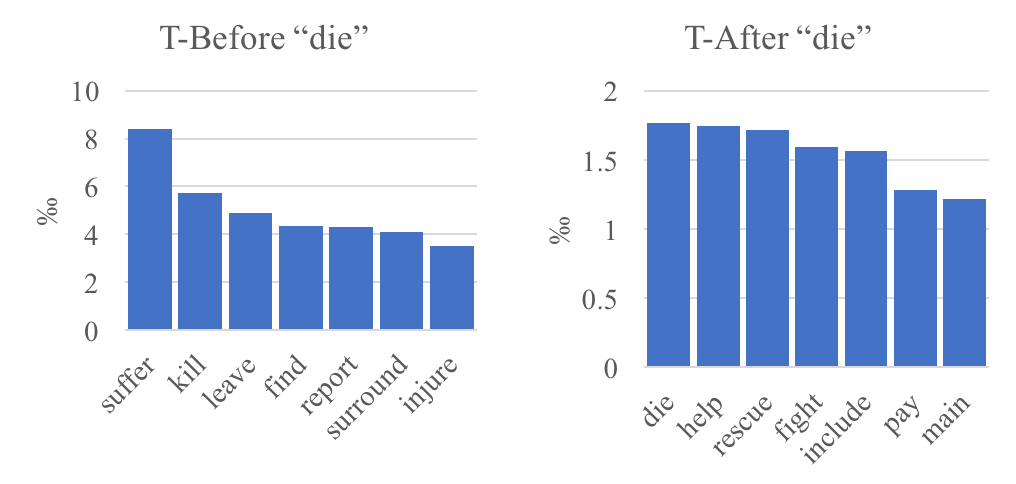}
		\caption{die}
	\end{subfigure}
	~
	\begin{subfigure}[htbp!]{0.49\textwidth}
		\centering
		\includegraphics[width=\textwidth]{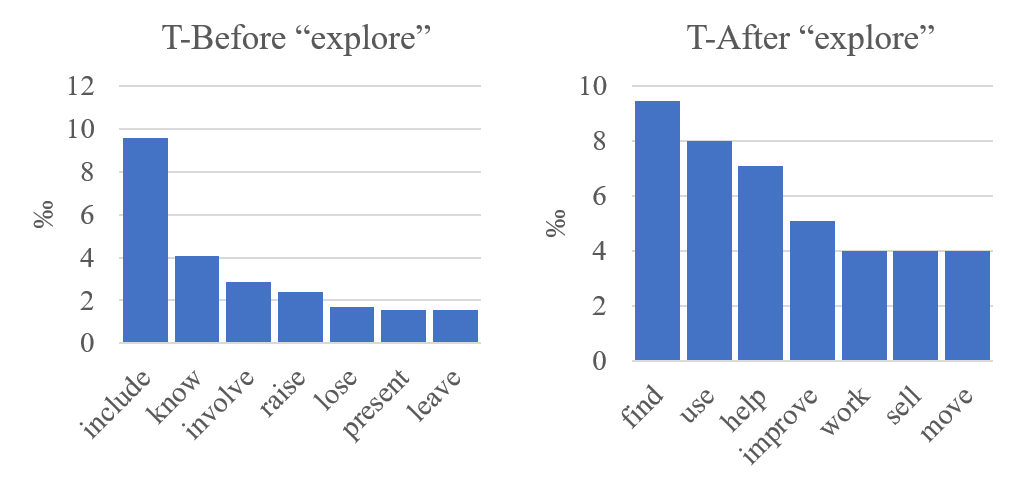}
		\caption{explore}
	\end{subfigure}
	
	\begin{subfigure}[htbp!]{0.49\textwidth}
		\centering
		\includegraphics[width=\textwidth]{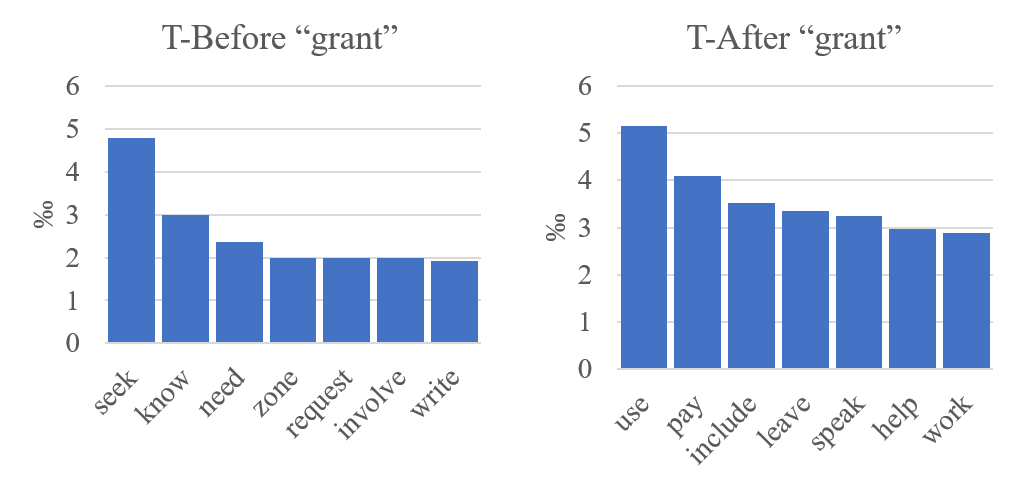}
		\caption{grant}
	\end{subfigure}
	~
	\begin{subfigure}[htbp!]{0.49\textwidth}
		\centering
		\includegraphics[width=\textwidth]{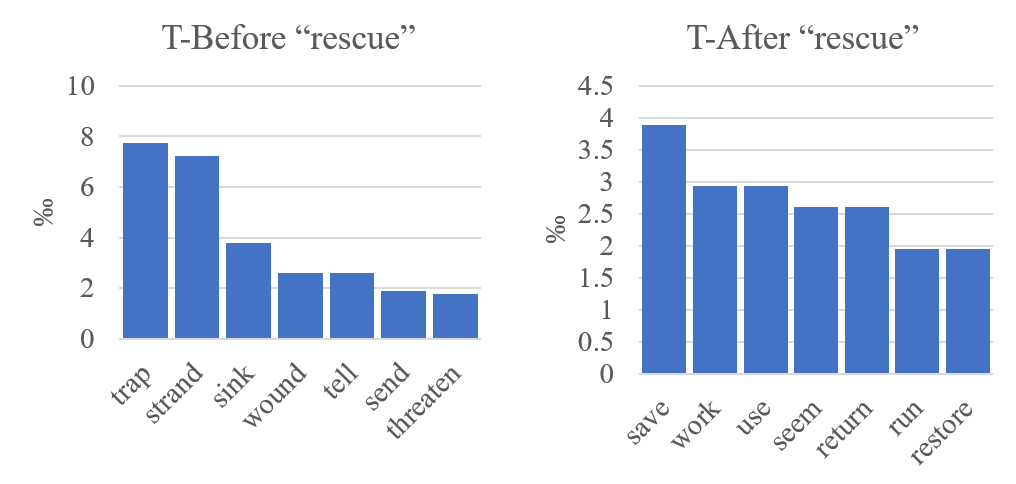}
		\caption{rescue}
	\end{subfigure}
	\caption{\small \textbf{A few more examples in addition to Fig.~\ref{fig:distribution}.}}
	\label{fig:app:distribution}
\end{figure*}

In Sec.~\ref{subsubsec:distribution}, we have shown some interesting examples with events that temporally precede or follow a certain event.
In addition to those examples shown in Fig.~\ref{fig:distribution}, it is actually very easy to find more interesting examples. We have picked some and shown them in Fig.~\ref{fig:app:distribution}. Again, there are potential errors in these figures, but the overall quality is intuitively appealing.

\begin{figure*}
	\centering
	\includegraphics[width=.9\textwidth]{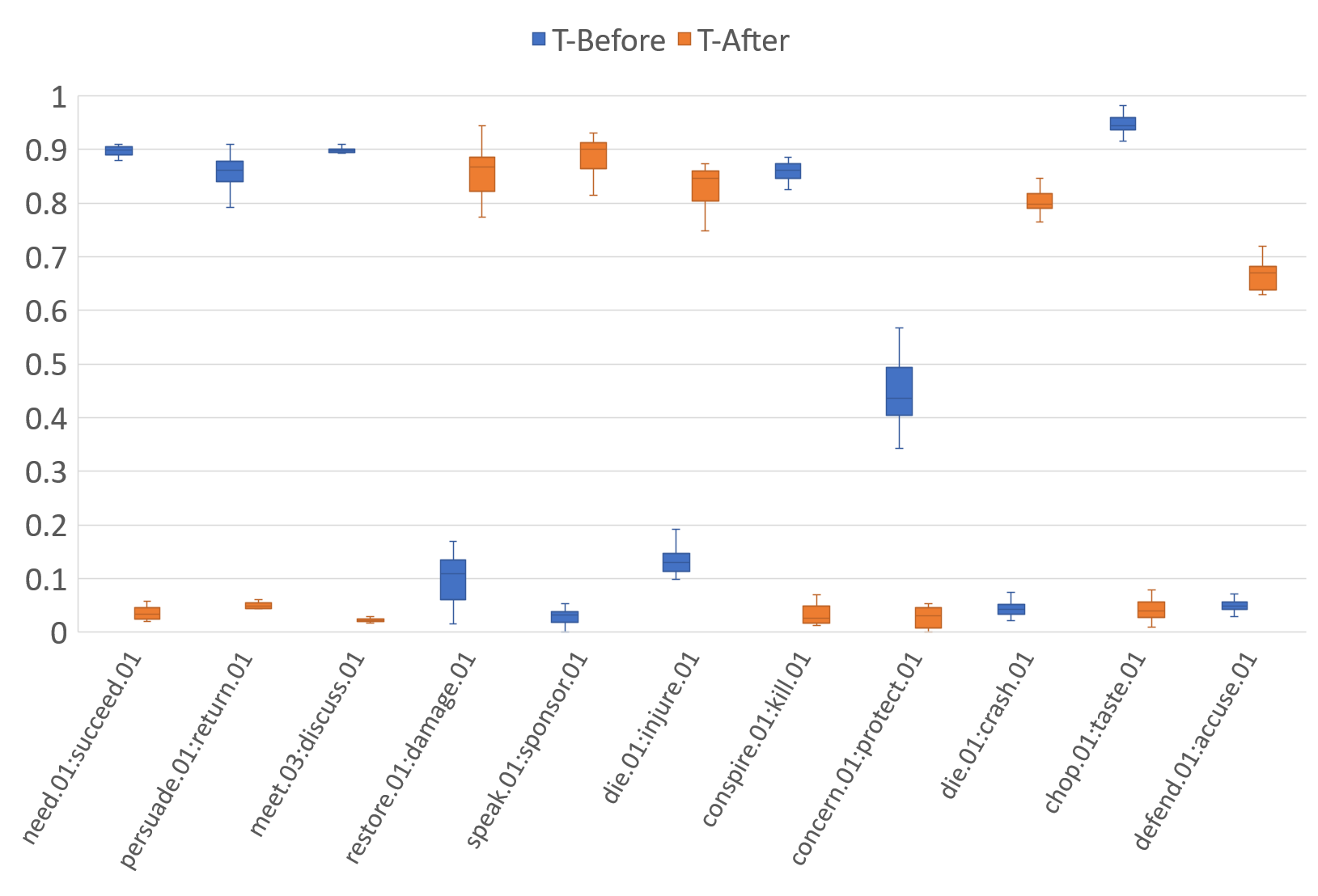}
	\caption{\textbf{Confidence intervals} from 10-fold bootstrapping.}
	\label{fig:app:conf}
\end{figure*}

In Sec.~\ref{subsec:quality}, we have shown the quality analysis of \ourkb{}~in Tables~\ref{tab:thresholding}-\ref{tab:causal} via the TimeBank-Dense dataset \cite{CassidyMcChBe14} and the EventCausality dataset \cite{DoChRo11}.
Here we further provide confidence level analysis on some randomly selected event pairs.
Specifically, we performed a 10-fold bootstrapping. In each fold, we randomly selected 50\% of the graphs in \ourkb{}~(i.e., from $\{G_i\}_{i=1}^N$) with replacement.
Then we re-calculated the prior statistics $\{f_r(v_i,v_j)\}$ (Eq.~\eqref{eq:all label prior}) in each fold.
By considering $\{f_r(v_i,v_j)\}$ as a random variable, we now obtained 10 realizations of it, so we can evaluate the confidence level.
In Fig.~\ref{fig:app:conf} (with color), we show the confidence levels of several randomly selected pairs of events.
From Fig.~\ref{fig:app:conf}, we can see that the prior statistics are actually rather stable, indicating that the prior statistics represented by \ourkb{}~is indeed a notion of knowledge underlying natural language text.

\end{document}